\documentclass[10pt,twocolumn,letterpaper]{article}

\usepackage{cvpr}
\usepackage{times}
\usepackage{epsfig}
\usepackage{graphicx}
\usepackage{amsmath}
\usepackage{amssymb}
\usepackage{multirow}
\usepackage{algorithm}
\usepackage{algpseudocode}

\usepackage[pagebackref=true,breaklinks=true,letterpaper=true,colorlinks,bookmarks=false]{hyperref}

\cvprfinalcopy 

\def\cvprPaperID{5562} 

\ifcvprfinal\fi
\begin{document}

\title{Learning Disentangled Representations of Satellite Image Time Series}

\author{\textbf{Eduardo H. Sanchez}\\
IRT Saint Exup\'{e}ry, IRIT\\
Toulouse, France\\
\and
\textbf{Mathieu Serrurier}\\
IRIT\\
Toulouse, France\\
\and
\textbf{Mathias Ortner}\\
IRT Saint Exup\'{e}ry\\
Toulouse, France\\
}

\maketitle

\begin{abstract}

In this paper, we investigate how to learn a suitable representation of satellite image time series in an unsupervised manner by leveraging large amounts of unlabeled data. Additionally, we aim to disentangle the representation of time series into two representations: a shared representation that captures the common information between the images of a time series and an exclusive representation that contains the specific information of each image of the time series. To address these issues, we propose a model that combines a novel component called cross-domain autoencoders with the variational autoencoder (VAE) and generative adversarial network (GAN) methods. In order to learn disentangled representations of time series, our model learns the multimodal image-to-image translation task. We train our model using satellite image time series from the Sentinel-2 mission. Several experiments are carried out to evaluate the obtained representations. We show that these disentangled representations can be very useful to perform multiple tasks such as image classification, image retrieval, image segmentation and change detection.

\end{abstract}

\section{Introduction}

Deep learning has demonstrated impressive performance on a variety of tasks such as image classification, object detection, semantic segmentation, among others. Typically, these models create internal abstract representations from raw data in a supervised manner. Nevertheless, supervised learning is a limited approach since it requires large amounts of labeled data. It is not always possible to obtain labeled data since it requires time, effort and resources. As a consequence, semi-supervised or unsupervised algorithms have been developed to reduce the required number of labels. Unsupervised learning is intended to learn useful representations of data easily transferable for further usage. As using smart data representations is important, another desirable property of unsupervised methods is to perform dimensionality reduction while keeping the most important characteristics of data. Classical methods are principal component analysis (PCA) or matrix factorization. For the same purpose, autoencoders learn to compress data into a low-dimensional representation and then, to uncompress that representation into the original data. An autoencoder variant is the variational autoencoder (VAE) introduced by Kingma and Welling \cite{Kingma2013} where the low-dimensional representation is constrained to follow a prior distribution. The VAE provides a way to extract a low-dimensional representation while learning the probability distribution of data. Other unsupervised methods of learning the probability data distribution have been recently proposed using generative models. A generative model of particular interest is generative adversarial networks (GANs) introduced by Goodfelow \etal \cite{Goodfellow2014, Goodfellow2016}.

In this work, we present a model that combines the VAE and GAN methods in order to create a useful representation of satellite image time series in an unsupervised manner. To create these representations we propose to learn the image-to-image translation task introduced by Isola \etal \cite{isola2017image} and Zhu \etal \cite{Zhu2017}. Given two images from a time series, we aim to translate one image into the other one. Since both images are acquired at different times, the model should learn the common information between these images as well as their differences to perform translation. We also aim to create a disentangled representation into a shared representation that captures the common information between the images of a time series and an exclusive representation that contains the specific information of each image. For instance, the knowledge about the specific information of each image could be useful to perform change detection.

Since we aim to generate any image of the time series from any of its images, we address the problem of multimodal generation, \ie multiple output images can be generated from a single input image. For instance, an image containing harvested fields could be translated into an image containing growing crop fields, harvested fields or a combination of both. 

Our approach is inspired by the BicycleGAN model introduced by Zhu \etal \cite{zhu2017toward} to address multimodal generation and the model presented by Gonzalez-Garcia \etal \cite{gonzalez2018image} to address representation disentanglement. 

In this work, the following contributions are made. First, we propose a model that combines the cross-domain autoencoder principle proposed by Gonzalez-Garcia \etal \cite{gonzalez2018image} under the GAN and VAE constraints to address representation disentanglement and multimodal generation. Our model is adapted to satellite image time series analysis using a simple architecture. Second, we show that our model is capable to process a huge volume of high-dimensional data such as Sentinel-2 image time series in order to create feature representations. Third, our model generates a disentangled representation that isolates the common information of the entire time series and the exclusive information of each image. Finally, our experiments suggest that these feature representations are useful to perform several tasks such as image classification, image retrieval, image segmentation and change detection.


\section{Related work}

\textbf{Variational autoencoder (VAE)}. In order to estimate the data distribution of a dataset, a common approach is to maximize the log-likelihood function given the samples of the dataset. A lower bound of the log-likelihood is introduced by Kingma and Welling \cite{Kingma2013}. To learn the data distribution, the authors propose to maximize the lower bound instead of the log-likelihood function which in some cases is intractable. The model is implemented using an autoencoder architecture and trained via a stochastic gradient descent method. It is an interesting method since it creates a low-dimensional representation where relevant attributes of data are captured.

\textbf{Generative adversarial networks (GAN)}. Due to its great success in many different domains, GANs \cite{ Goodfellow2014, Goodfellow2016} have become one of the most important research topics. The GAN model can be thought of as a game between two players: the generator and the discriminator. In this setting, the generator aims to produce samples that look like drawn from the same distribution as the training samples. On the other hand, the discriminator receives samples to determine whether they are real (dataset samples) or fake (generated samples). The generator is trained to fool the discriminator by learning a mapping function from a latent space which follows a prior distribution to the data space. However, traditional GANs (DCGAN \cite{radford2015unsupervised}, LSGAN \cite{Mao2017}, BEGAN \cite{Berthelot2017}, WGAN \cite{arjovsky2017wasserstein}, WGAN-GP \cite{Gulrajani2017}, LaplacianGAN \cite{Denton2015}, EBGAN \cite{zhao2016energy}, among others) does not provide a means to learn the inverse mapping from the data space to the latent space. To solve this problem, several models were proposed such us BiGAN \cite{donahue2016adversarial} or VAE-GAN \cite{pmlr-v48-larsen16} which include an encoder from the data space to the latent space in the model. The data representation obtained in the latent space via the encoder can be used for other tasks as shown by Donahue \etal \cite{donahue2016adversarial}. 

\textbf{Image-to-image translation}. It is one of the most popular applications using conditional GANs \cite{Mirza2014}. The image-to-image translation task consists of learning a mapping function between an input image domain and an output image domain. Impressive results have been achieved by the pix2pix \cite{isola2017image} and cycleGAN \cite{Zhu2017} models. Nevertheless, most of these models are monomodal. That is, there is a unique output image for a given input image.

\textbf{Multimodal image-to-image translation}. One of the limitations of previous models is the lack of diversity of generated images. Certain models address this problem by combining the GAN and VAE methods. On the one hand, GANs are used to generate realistic images while VAE is used to provide diversity in the output domain. Recent work that deals with multimodal output is presented by Gonzalez-Garcia \etal \cite{gonzalez2018image}, Zhu \etal \cite{zhu2017toward}, Huang \etal \cite{huang2018multimodal}, Lee \etal \cite{lee2018stochastic} and Ma \etal \cite{ma2018exemplar}. In particular, to be able to generate an entire time series from a single image, we adopt the principle of the BicycleGAN model proposed by Zhu \etal \cite{zhu2017toward} where a low-dimensional latent vector represents the diversity of the output domain. However, while the BicycleGAN model is mainly focus on image generation, we only consider the image-to-image translation task as a way to learn suitable feature representations. For image generation purpose, the output diversity is conditioned at the encoder input level in the BicycleGAN model. Instead the output diversity is conditioned at the decoder input level in our model.

\textbf{Disentangled feature representation}. Recent work is focused on learning disentangled representations by isolating the factors of variation of high-dimensional data in an unsupervised manner. A disentangled representation can be very useful for several tasks that require knowledge of these factors of variation. Chen \etal \cite{Chen2016} propose an objective function based on the maximization of the mutual information. Gonzalez-Garcia \etal \cite{gonzalez2018image} propose a model based on VAE-GAN image translators and a novel network component called cross-domain autoencoders. This model separates the feature representation of two image domains into three parts: the shared part which contains common information from both domains and the exclusive parts which only contain factors of variation that are specific to each domain. We propose a model that combines the cross-domain autoencoder component under the VAE and GAN constraints in order to create representations containing the common information of the entire time series and the exclusive information of each image. The VAE is used to create a low-dimensional representation that encodes the image variations related to acquisition time and the GAN is used to evaluate the generated image at a given time. We introduce a model adapted for satellite image time series using a simpler architecture since only four functions (the shared representation encoder, the exclusive representation encoder, the decoder and the discriminator) must be learned.
 
\begin{figure*}[t]
\begin{center}
\includegraphics[width=0.6\linewidth]{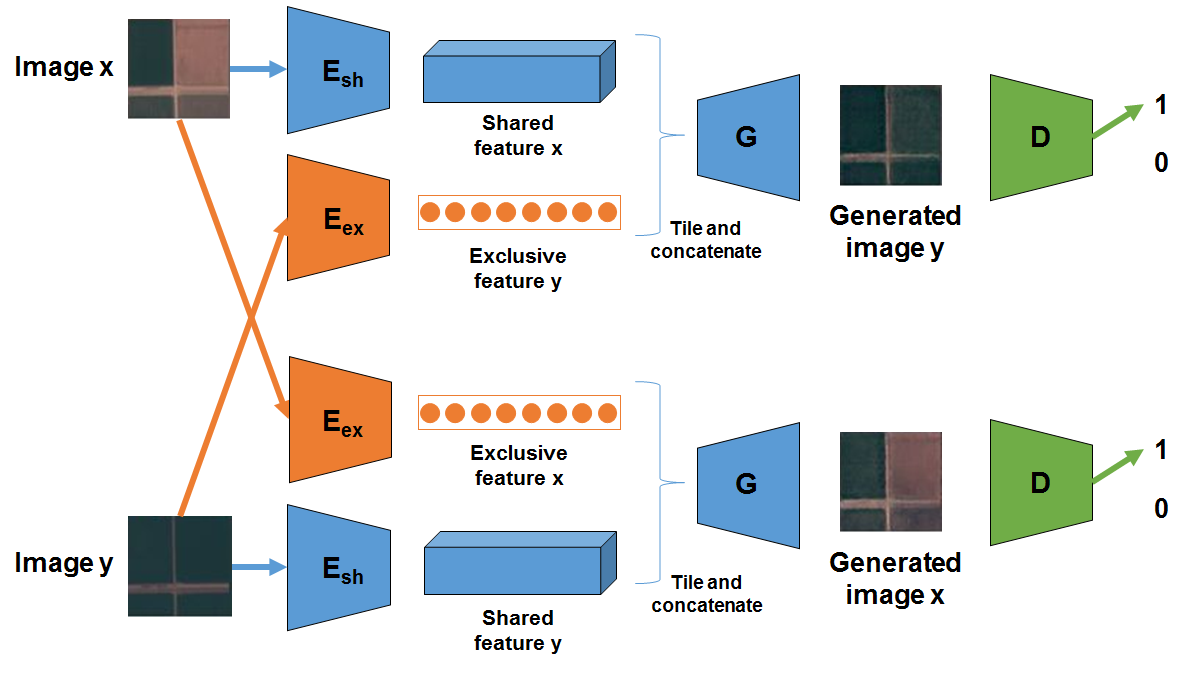}
\end{center}
\caption{Model overview. The model goal is to learn both image transitions: $\mathbf{x} \rightarrow \mathbf{y}$ and $\mathbf{y} \rightarrow \mathbf{x}$. Both images are passed through the network $\mathbf{E}_{sh}$ in order to extract their shared representations. On the other hand, the network $\mathbf{E}_{ex}$ extracts the exclusive representations corresponding to images $\mathbf{x}$ and $\mathbf{y}$. The exclusive representation encoder output is constrained to follow a standard normal distribution. In order to generate the image $\mathbf{y}$, the decoder network $\mathbf{G}$ takes the shared feature of image $\mathbf{x}$ and the exclusive feature of image $\mathbf{y}$. A similar procedure is performed to generate the image $\mathbf{x}$. Finally, the discriminator $\mathbf{D}$ is used to evaluate the generated images.}
\label{fig:ModelOverview}
\end{figure*}

\section{Method}

Let $\mathbf{x}, \mathbf{y}$ be two images randomly sampled from a given time series $\mathbf{t}$ in a region $\mathbf{c}$. Let $\mathcal{X}$ be the image domain where these images belong to and let $\mathcal{R}$ be the representation domain of these images. The representation domain $\mathcal{R}$ is divided into two subdomains $\mathcal{S}$ and $\mathcal{E}$, $\mathcal{R} = \left[\mathcal{S}, \mathcal{E}\right]$. The subdomain $\mathcal{S}$ contains the common information between images $\mathbf{x}$ and $\mathbf{y}$ and the subdomain $\mathcal{E}$ contains the particular information of each image. Since images $\mathbf{x}$ and $\mathbf{y}$ belong to the same time series, their shared representations must be identical, \ie $\mathcal{S}_{\mathbf{x}} = \mathcal{S}_{\mathbf{y}}$. On the other hand, as images are acquired at different times, their exclusive representations $\mathcal{E}_{\mathbf{x}}$ and $\mathcal{E}_{\mathbf{y}}$ correspond to the specific information of each image.

We propose a model that learns the transition from $\mathbf{x}$ to $\mathbf{y}$ as well as the inverse transition from $\mathbf{y}$ to $\mathbf{x}$. In order to accomplish this, an autoencoder-like architecture is used. In Figure \ref{fig:ModelOverview}, an overview of the model can be observed. Let $\mathbf{E}_{sh}:\mathcal{X} \rightarrow \mathcal{S}$ be the shared representation encoder and $\mathbf{E}_{ex}:\mathcal{X} \rightarrow \mathcal{E}$ be the exclusive representation encoder. To generate the image $\mathbf{y}$, the shared feature of $\mathbf{x}$, \ie $\mathbf{E}_{sh}(\mathbf{x})$, and the exclusive feature of $\mathbf{y}$, \ie $\mathbf{E}_{ex}(\mathbf{y})$ are computed. Then both representations are passed through the decoder function $\mathbf{G}:\mathcal{R} \rightarrow \mathcal{X}$ which generates a reconstructed image $\mathbf{G}(\mathbf{E}_{sh}(\mathbf{x}), \mathbf{E}_{ex}(\mathbf{y}))$. A similar process is followed to reconstruct the image $\mathbf{x}$. Then, these images are passed through a discriminator function $\mathbf{D}:\mathcal{X} \rightarrow \left[0,1\right]$ in order to evaluate the generated images. 

The model functions $\mathbf{E}_{ex}$, $\mathbf{E}_{sh}$, $\mathbf{G}$ and $\mathbf{D}$ are represented by neural networks with parameters $\theta_{\mathbf{E}_{ex}}$, $\theta_{\mathbf{E}_{sh}}$ and $\theta_{\mathbf{G}}$ and $\theta_{\mathbf{D}}$, respectively. The training procedure to learn these parameters is explained below.

\subsection{Objective function}

As the work presented by Zhu \etal \cite{zhu2017toward} and Gonzalez-Garcia \etal \cite{gonzalez2018image}, our objective function is composed of several terms in order to obtain a disentangled representation.

Concerning the shared representation, images $\mathbf{x}$ and $\mathbf{y}$ must have identical shared feature representations, \ie $\mathbf{E}_{sh}(\mathbf{x}) = \mathbf{E}_{sh}(\mathbf{y})$. A simple solution is to minimize the $L_1$ distance between their shared feature representations as can be seen in Equation \ref{eq:sharedloss}.
\begin{equation}\label{eq:sharedloss}
\mathbf{L}_{1}^{sh} = \mathbb{E}_{\mathbf{x}, \mathbf{y} \sim \mathcal{X}} \left[ \lvert \mathbf{E}_{sh}(\mathbf{x}) - \mathbf{E}_{sh}(\mathbf{y}) \rvert \right]
\end{equation}

The exclusive representation must only contain the particular information that corresponds to each image. To enforce the disentanglement between shared and exclusive features, we include a reconstruction loss in the objective function where the shared representations of $\mathbf{x}$ and $\mathbf{y}$ are switched. The loss term corresponding to the reconstruction of image $\mathbf{x}$ is represented in Equation \ref{eq:vaereconstructionloss_x}. Moreover, this loss term can be thought of as the reconstruction loss in the VAE model \cite{Kingma2013} which maximizes a lower bound of the log-likelihood function. As we enforce representation disentanglement, we simultaneously maximize the log-likelihood function which is equivalent to Kullback-Leibler divergence minimization between the real distribution and the generated distribution.
\begin{equation} 
\mathbf{L}_{1}^{\mathbf{x}, \mathbf{y}} = \mathbb{E}_{\mathbf{x}, \mathbf{y} \sim \mathcal{X}} \left[ \lvert \mathbf{x} - \mathbf{G}(\mathbf{E}_{sh}(\mathbf{y}), \mathbf{E}_{ex}(\mathbf{x})) \rvert \right] \label{eq:vaereconstructionloss_x}
\end{equation}

On the other hand, the lower bound proposed in the VAE model constraints the feature representation to follow a prior distribution. In our model, we only force the exclusive feature representation to be distributed as a standard normal distribution $\mathcal{N}(0, I)$ in order to generate multiple outputs by sampling from this space during inference. In contrast to the approach employed by Gonzalez-Garcia \etal \cite{gonzalez2018image} which uses a GAN approach to constraint the exclusive feature distribution, a simpler solution which proves to be effective is to include a Kullback-Leibler divergence term between the distribution of the exclusive feature representation and the prior $\mathcal{N}(0, I)$. Assuming that the exclusive feature encoder $\mathbf{E}_{ex}(\mathbf{x})$ is distributed as a normal distribution $\mathcal{N}(\mu_{\mathbf{E}_{ex}(\mathbf{x})}, \sigma_{\mathbf{E}_{ex}(\mathbf{x})})$, the Kullback-Leibler divergence can be written as follows
\begin{equation}\label{eq:kullbackleiblerdistance}
\resizebox{.9\hsize}{!}{$\mathbf{L}_{KL}^{\mathbf{x}} = -\frac{1}{2} \mathbb{E}_{\mathbf{x} \sim \mathcal{X}} \left[ 1 + \log(\sigma_{\mathbf{E}_{ex}(\mathbf{x})}^2) - \mu_{\mathbf{E}_{ex}(\mathbf{x})}^2 - \sigma_{\mathbf{E}_{ex}(\mathbf{x})}^2 \right]$}
\end{equation}

In order to minimize the distance between the real distribution and the generated distribution of images, a LSGAN loss \cite{Mao2017} is included in the objective function. The discriminator is trained to maximize the probability of assigning the correct label to real images and generated images while the generator is trained to fool the discriminator by classifying generated images as real, \ie $\mathbf{D}(\mathbf{G}(\mathbf{E}_{sh}(\mathbf{y}), \mathbf{E}_{ex}(\mathbf{x}))) \rightarrow 1$. The corresponding loss term for image $\mathbf{x}$ and its reconstructed version can be seen in Equation \ref{eq:gan_x} where the discriminator maximizes this term while the generator minimizes it.
\begin{align}
\begin{split}
\mathbf{L}_{GAN}^{\mathbf{x}} =& \ \mathbb{E}_{\mathbf{x} \sim \mathcal{X}} \left[ \left(\mathbf{D}(\mathbf{x})\right)^{2} \right] + \\[0.20em]
& \ \mathbb{E}_{\mathbf{x}, \mathbf{y} \sim \mathcal{X}} \left[ \left(1 - \mathbf{D}(\mathbf{G}(\mathbf{E}_{sh}(\mathbf{y}), \mathbf{E}_{ex}(\mathbf{x})))\right)^{2} \right]
\end{split}\label{eq:gan_x}
\end{align}

To summarize, the training procedure can be seen as a minimax game (Equation \ref{eq:totallosses}) where the objective function $\mathcal{L}$ is minimized by the generator functions of the model ($\mathbf{E}_{ex}$, $\mathbf{E}_{sh}$, $\mathbf{G}$) while it is maximized by the discriminator $\mathbf{D}$.
\begin{equation}\label{eq:totallosses}
\begin{split}
\min_{\mathbf{E}_{ex}, \mathbf{E}_{sh}, \mathbf{G}} \max_{\mathbf{D}} \ \mathcal{L} =& \ \mathbf{L}_{GAN}^{\mathbf{x}} + \mathbf{L}_{GAN}^{\mathbf{y}} + \lambda_{L_1} \left(\mathbf{L}_{1}^{\mathbf{x}, \mathbf{y}} + \mathbf{L}_{1}^{\mathbf{y}, \mathbf{x}}\right)\\[0.25em]
 & + \lambda_{L_{KL}} (\mathbf{L}_{KL}^{\mathbf{x}} + \mathbf{L}_{KL}^{\mathbf{y}}) + \lambda_{L_1}^{sh} \mathbf{L}_{1}^{sh}
\end{split}
\end{equation}

Where $\lambda_{L_1}$, $\lambda_{L_1}^{sh}$ and $\lambda_{L_{KL}}$ are constant coefficients to weight the loss terms.

\subsection{Implementation}

\textbf{Network architectures}: Our model is architectured around four network blocks: the shared feature encoder, the exclusive feature encoder, the decoder and the discriminator. The shared feature encoder is composed of 5 convolutional layers while the exclusive feature encoder is composed of a first convolutional layer and three consecutive ResNet blocks \cite{He2016a}. Since the exclusive feature encoder must provide a normally distributed vector, 2 fully-connected layers of size 64 are appended on top of the ResNet blocks to estimate its mean value $\mu$ and standard deviation $\sigma$. The decoder consists of 4 transposed convolutional layers. Finally, the discriminator is composed of 5 convolutional layers. During training and test experiments, we use batch normalization in all the networks. All the layers use a kernel of size $4 \times 4$, a stride of 2 and leaky ReLU as activation function (except the discriminator and the decoder outputs where the sigmoid and hyperbolic tangent functions are used, respectively).

\textbf{Optimization setting}: To train our model, we use batches of 64 randomly selected pairs of images of size $64 \times 64 \times 4$ from our time series dataset. Every network is trained from scratch by using randomly initialized weights as starting point. The learning rate is implemented as a staircase function which starts with an initial value of 0.0002 and decays every 50000 iterations. We use Adam optimizer to update the network weights using a $\beta = 0.5$ during 150000 iterations. Concerning the loss coefficients, we use the following values: $\lambda_{L_1} = 10$, $\lambda_{L_1}^{sh} = 0.5$ and $\lambda_{L_{KL}} = 0.01$ during training. The training procedure is summarized in Algorithm \ref{alg:trainingalgorithm}.

\begin{algorithm}
\caption{Training algorithm.}
\label{alg:trainingalgorithm}
  \begin{algorithmic}[1]
			\State Random initialization of model parameters \\ $(\theta_{\mathbf{D}}^{(0)}, \theta_{\mathbf{E}_{sh}}^{(0)}, \theta_{\mathbf{E}_{ex}}^{(0)}, \theta_{\mathbf{G}}^{(0)})$
      \For{$k=1$; $k= k + 1$; $k<\text{number of iterations}$}
				\State Sample a batch of  $m$ time series $\{\mathbf{t}_{s}^{(1)},...,\mathbf{t}_{s}^{(m)}\}$
				\State Sample a batch of  $m$ image pairs  \\ \hspace{0.5cm} $\{ (\mathbf{x}^{(1)}, \mathbf{y}^{(1)}),...,(\mathbf{x}^{(m)}, \mathbf{y}^{(m)})\}$ from $\{\mathbf{t}_{s}^{(i)}\}$ 
				\State Compute $\mathcal{L}^{(k)}(\mathbf{x}^{(i)}, \mathbf{y}^{(i)}, \theta_{\mathbf{D}}^{(k)}, \theta_{\mathbf{E}_{sh}}^{(k)}, \theta_{\mathbf{E}_{ex}}^{(k)}, \theta_{\mathbf{G}}^{(k)})$
			  \begin{equation}\label{eq:loss_estimated} 
				\begin{split}
				\mathcal{L}^{(k)} =& \ \frac{1}{m} \sum_{i=1}^{m} \left[ \left(\mathbf{D}(\mathbf{x}^{(i)})\right)^{2} + \left(\mathbf{D}(\mathbf{y}^{(i)})\right)^{2} \right. \\
				& + \left(1 - \mathbf{D}(\mathbf{G}(\mathbf{E}_{sh}(\mathbf{y}^{(i)}), \mathbf{E}_{ex}(\mathbf{x}^{(i)})))\right)^{2}\\
				& + \left(1 - \mathbf{D}(\mathbf{G}(\mathbf{E}_{sh}(\mathbf{x}^{(i)}), \mathbf{E}_{ex}(\mathbf{y}^{(i)})))\right)^{2} \\
				& + \lambda_{L_1} \left( \lvert \mathbf{x}^{(i)} - \mathbf{G}(\mathbf{E}_{sh}(\mathbf{y}^{(i)}), \mathbf{E}_{ex}(\mathbf{x}^{(i)})) \rvert \right. \\
				& + \left. \lvert \mathbf{y}^{(i)} - \mathbf{G}(\mathbf{E}_{sh}(\mathbf{x}^{(i)}), \mathbf{E}_{ex}(\mathbf{y}^{(i)})) \rvert \right) \\
				& + \lambda_{L_1}^{sh} \left( \lvert \mathbf{E}_{sh}(\mathbf{x}^{(i)}) - \mathbf{E}_{sh}(\mathbf{y}^{(i)}) \rvert \right) \\
				& - \frac{1}{2} \lambda_{L_{KL}} \left(2 + \log(\sigma_{\mathbf{E}_{ex}(\mathbf{x}^{(i)})}^2) - \mu_{\mathbf{E}_{ex}(\mathbf{x}^{(i)})}^2  \right. \\
				& - \sigma_{\mathbf{E}_{ex}(\mathbf{x}^{(i)})}^2 + \log(\sigma_{\mathbf{E}_{ex}(\mathbf{y}^{(i)})}^2) - \mu_{\mathbf{E}_{ex}(\mathbf{y}^{(i)})}^2 \\
				& \left. \left. - \sigma_{\mathbf{E}_{ex}(\mathbf{y}^{(i)})}^2 \right) \right]
				\end{split}
				\end{equation}
				\State Update the model parameters:
				\begin{align}\label{eq:model_parameters}
				\theta_{\mathbf{D}}^{(k+1)} &\leftarrow \text{Adam} \left(-\nabla_{\theta_{\mathbf{D}}^{(k)}}\mathcal{L}^{(k)}, \theta_{\mathbf{D}}^{(k)}\right)\\
				\theta_{\mathbf{E}_{sh}}^{(k+1)} &\leftarrow \text{Adam} \left(\nabla_{\theta_{\mathbf{E}_{sh}}^{(k)}}\mathcal{L}^{(k)}, \theta_{\mathbf{E}_{sh}}^{(k)} \right)\\
				\theta_{\mathbf{E}_{ex}}^{(k+1)} &\leftarrow \text{Adam} \left(\nabla_{\theta_{\mathbf{E}_{ex}}^{(k)}}\mathcal{L}^{(k)}, \theta_{\mathbf{E}_{ex}}^{(k)}\right)\\
				\theta_{\mathbf{G}}^{(k+1)} &\leftarrow \text{Adam} \left(\nabla_{\theta_{\mathbf{G}}^{(k)}}\mathcal{L}^{(k)}, \theta_{\mathbf{G}}^{(k)}\right)
				\end{align}
      \EndFor
  \end{algorithmic}
\end{algorithm}

\section{Experiments}

\subsection{Sentinel-2}

The Sentinel-2 mission is composed of a constellation of 2 satellites that orbit around the Earth providing an entire Earth coverage every 5 days. Both satellites acquire images at 13 spectral bands using different spatial resolutions. In this paper, we use the RGBI bands which correspond to bands at 10m spatial resolution. In order to organize the data acquired by the mission, Earth surface is divided into square tiles of approximately 100 km on each side. One tile acquired at a particular time is referred to as a granule.

To create our dataset, we selected 42 tiles containing several regions of interest such as the Amazon rainforest, the Dead Sea, the city of Los Angeles, the Great Sandy Desert, circular fields in Saudi Arabia, among others. As explained by Kempeneers and Soille~\cite{kempeneers2017optimizing}, many of the acquired granules might carry useless information. In our case, the availability of granules for a given tile depends on two factors: the cloud coverage and the image completeness. Therefore, we defined a threshold in order to avoid these kind of problems that affect Earth observation by setting a cloud coverage tolerance of 2\% and completeness tolerance of 85\%. For each tile, we extracted 12 granules from March 2016 to April 2018. Then, we selected 25 patches of size $1024 \times 1024$ from the center of the tiles to reduce the effect of the satellite orbit view angle. Finally, our dataset is composed of 1050 times series each of which is composed of 12 images of size $1024 \times 1024 \times 4$.

\begin{figure}[t]
\begin{center}
\includegraphics[width=0.9\linewidth]{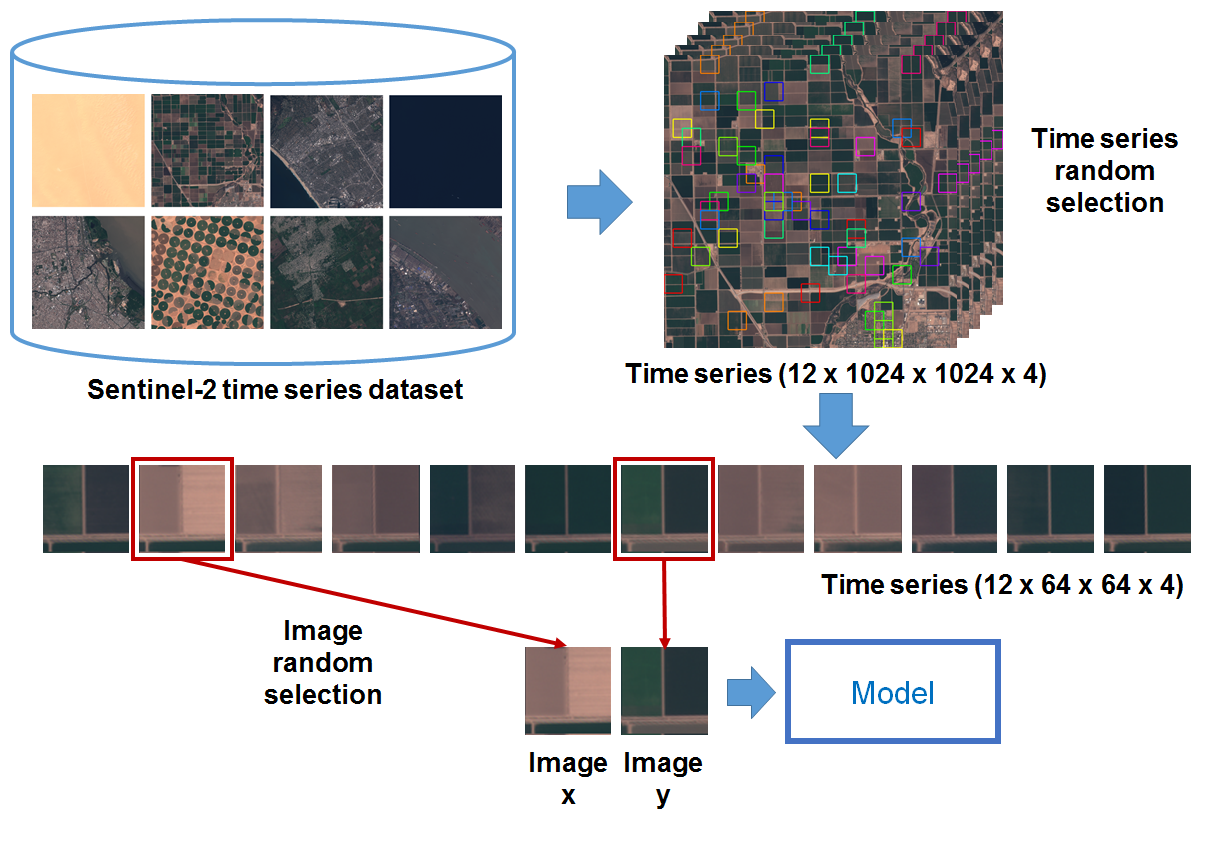}
\end{center}
\caption{Training data selection. A batch of smaller time series is randomly sampled from the dataset. At each iteration two images are randomly selected from each time series to be used as input for our model.}
\label{fig:TimeSeriesDataSelection}
\end{figure}

In order to analyze the entire time series using smaller patches the following strategy is applied: a batch of time series composed of images of size $64 \times 64 \times 4$ is randomly sampled from the time series of size $1024 \times 1024 \times 4$. Since our model takes 2 images as input, at each iteration two images are randomly selected from the time series to be used as input for our model. Thus, the whole time series is learned as the training procedure progresses. Data sampling procedure is depicted in Figure \ref{fig:TimeSeriesDataSelection}.  

To evaluate the model performance and the learned representations, we perform several supervised and unsupervised experiments on Sentinel-2 data as suggested by Theis in \cite{theis2015note}. We evaluate our model on: a) image-to-image translation to validate the representation disentanglement; b) image retrieval, image classification and image segmentation to validate the shared representation and c) change detection to analyze the exclusive representation.

\subsection{Image-to-image translation}

\begin{figure*}[t]
\begin{center}
\begin{tabular}{c c c} 	
\raisebox{-0.5\totalheight}{\includegraphics[width=55mm, height=55mm]{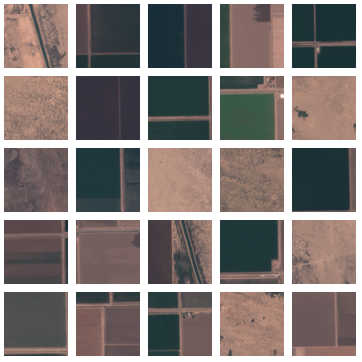}} & \raisebox{-0.5\totalheight}{\includegraphics[width=55mm, height=55mm]{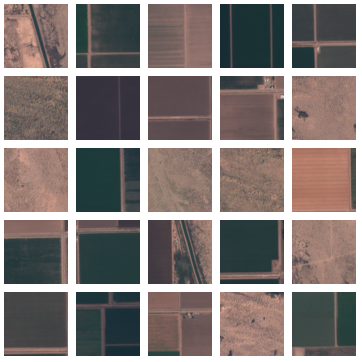}} & \raisebox{-0.5\totalheight}{\includegraphics[width=55mm, height=55mm]{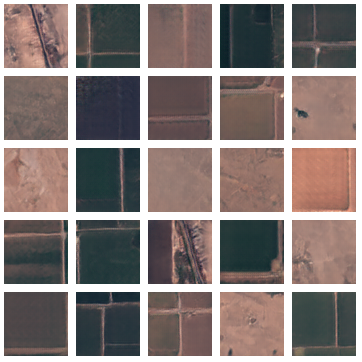}} \\[3.00cm]
(a) & (b) & (c)
\end{tabular}
\end{center}
\caption{Image translation performed on images of Brawley, California. (a) Images used to extract the shared features; (b) Images used to extract the exclusive features; (c) Generated images from the shared representation of (a) and the exclusive representation of (b).}
\label{fig:ImageTranslationExperiment0}
\end{figure*}

\begin{figure*}[h]
\begin{center}
\includegraphics[width=0.8\linewidth]{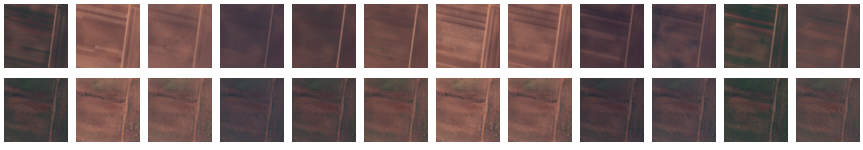}
\end{center}
\caption{Multimodal generation. The first row corresponds to a time series sampled from the test dataset. The second row corresponds to a time series where each image is generated by using the same shared feature and only modifying the exclusive feature.}
\label{fig:multimodal}
\end{figure*}

It seems natural to first test the model performance at image translation. We use a test dataset which is composed of time series acquired from different tiles to guarantee that training and test datasets are independent. For each dataset, 150 batches of 64 time series are randomly selected. It represents around 20k processed images of size $64 \times 64 \times 4$.

An example of image-to-image translation can be observed in Figure \ref{fig:ImageTranslationExperiment0}. For instance, let us consider the image in the third row, fifth column. The shared feature is extracted from an image $\mathbf{x}$ which corresponds to growing crop fields while the exclusive feature is extracted from another image $\mathbf{y}$ where fields have been harvested. Consequently, the generated image contains harvested fields which is defined by the exclusive feature of image $\mathbf{y}$. In general, generated images look realistic in both training and test datasets except for small details which are most likely due to the absence of skip connections in the generator part of the model.

We quantify the $L_1$ distance between generated images $\mathbf{G}(\mathbf{E}_{sh}(\mathbf{x}), \mathbf{E}_{ex}(\mathbf{y}))$ and images $\mathbf{y}$ used to extract the exclusive feature. Results can be observed in Table \ref{tab:ImageTranslationExperiment} (first row). Pixel values in generated images and real images are in the range of $[-1, 1]$, thus a mean difference of $0.0152$ in the training dataset indicates that the model performs well at image-to-image translation. A slightly difference is obtained in the test dataset where the $L_1$ distance is $0.0207$. 

A special image-to-image translation case is image autoencoding where the shared and exclusive features are extracted from the same image. Additionally, we compute the $L_1$ distance between images $\mathbf{x}$ and autoencoded images $\mathbf{G}(\mathbf{E}_{sh}(\mathbf{x}), \mathbf{E}_{ex}(\mathbf{x}))$ for comparison purpose in Table \ref{tab:ImageTranslationExperiment} (second row). Lower values in terms of $L_1$ distance are obtained with respect to those of image-to-image translation. It is important to note that input images are considerably well reconstructed even if this case is not considered during training. Finally, we perform times series reconstruction in order to show that the exclusive feature encodes the specific information of each image. An image is randomly selected from a time series to extract its shared feature. While keeping the shared feature constant and only modifying the exclusive feature, we reconstruct the original time series. Results in terms of $L_1$ distance between the original time series and the reconstructed one can be observed in Table \ref{tab:ImageTranslationExperiment} (third row). Similar values to those of image-to-image translation are obtained. An example of time series reconstruction can be seen in Figure \ref{fig:multimodal}.

\begin{table}[h]
\begin{center}
  \begin{tabular}{| c | c | c |}
    \hline
    Task & Training set & Test set \\ \hline
		\hline
    Image translation &  $0.0152 \pm 0.0573$ & $0.0207 \pm 0.0774$ \\ \hline
    Autoencoding      &  $0.0084 \pm 0.0309$ & $0.0087 \pm 0.0312$ \\ \hline
		Time series       &  $0.0177 \pm 0.0622$ & $0.0223 \pm 0.0828$ \\
    \hline
  \end{tabular}
\end{center}
\caption{Mean and standard deviation values of the $L_1$ distance for image-to-image translation (first row), image autoencoding (second row) and time series reconstruction (third row).}
\label{tab:ImageTranslationExperiment}
\end{table}

\subsection{Image retrieval}

\begin{figure}[!t]
\begin{center}
\begin{tabular}{c c} 	
\raisebox{-0.5\totalheight}{\includegraphics[width=40mm, height=40mm]{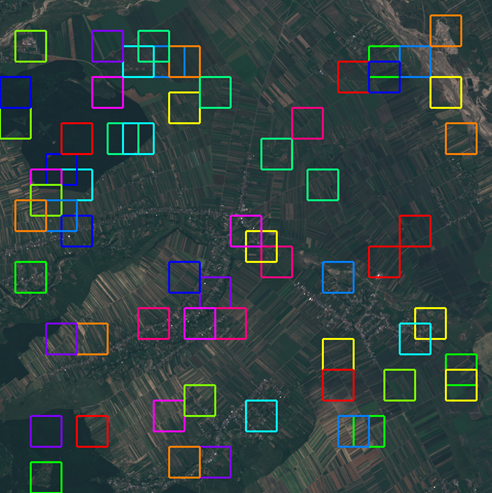}} & \raisebox{-0.5\totalheight}{\includegraphics[width=40mm, height=40mm]{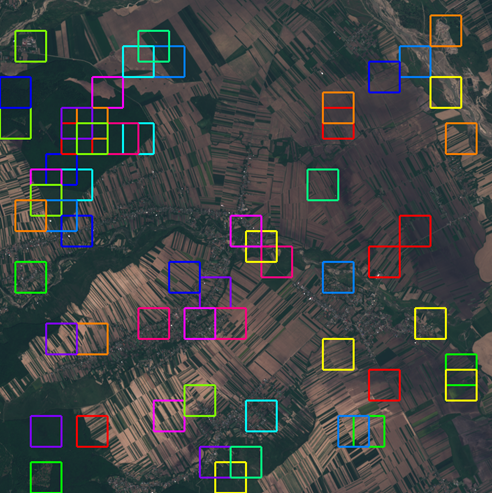}} \\[2.00cm]
(a) & (b)
\end{tabular}
\end{center}
\caption{Image retrieval using shared feature comparison. (a) Selected image from a time series where a batch of 64 patches (colored boxes) are extracted from; (b) Another image from the same time series is used to locate the selected patches. The algorithm plots colored boxes corresponding to the nearest patches in terms of shared feature distance.}
\label{fig:ImageRetrievalSharedFeatureBBPatches}
\end{figure}

In this experiment, we want to evaluate whether the shared feature provides information about the geographical location of time series via image retrieval. Given an image patch from a granule acquired at time $t_o$, we would like to locate it in a granule acquired at time $t_f$.
The procedure is the following: a time series of size $12 \times 1024 \times 1024 \times 4$ is randomly sampled from the dataset. Then, a batch of 64 image patches of size $64 \times 64 \times 4$ is randomly selected as shown in Figure \ref{fig:ImageRetrievalSharedFeatureBBPatches}(a). The corresponding shared features are extracted for each image of the batch. The main idea is to use the information provided by the shared feature to locate the image patches in every image of the time series. For each image of the time series, a sliding window of size $64 \times 64 \times 4$ is applied in order to explore the entire image. As the window slides, the shared features are extracted and compared to those of the images to be retrieved. The nearest image in terms of $L_1$ distance is selected as the retrieved image at each image of the time series. In our experiment, 150 time series of size $12 \times 1024 \times 1024 \times 4$ are analyzed. It represents around 115k images of size $64 \times 64 \times 4$ to be retrieved and 110M images of size $64 \times 64 \times 4$ to be analyzed.

To illustrate the retrieval algorithm, let us consider a test image of agricultural fields. We plot the patches to be retrieved in Figure \ref{fig:ImageRetrievalSharedFeatureBBPatches}(a) and the retrieved patches by the algorithm in Figure \ref{fig:ImageRetrievalSharedFeatureBBPatches}(b). As can be seen, even if some changes have occurred, the algorithm is able to spatially locate most of the patches. In spite of the seasonal changes in the agricultural fields, the algorithm performs correctly since the image retrieval leverages the shared representation which contains common information of the time series. Results in terms of Recall@1 are displayed in Table \ref{tab:ImageRetrievalSharedFeatures} (first row). We obtain a high value in terms of Recall@1 even if it is not so close to 1. This result can be explained since the dataset contains several time series from the desert, forest and ocean tiles which could be notoriously difficult to retrieve even for humans. For instance, image retrieval performs better in urban scenarios since the city provides details that can be easily identified in contrast to agricultural fields where distinguishing textures can be confusing. 

\begin{table}[h]
\begin{center}
\begin{tabular}{| c | c |}
    \hline
    Method &  Recall@1 \\ \hline
		\hline
    Shared features & 0.7372 \\ \hline
    Raw pixels & 0.5083 \\
    \hline
  \end{tabular}
\end{center}
\caption{Image retrieval results in terms of Recall@1  using the shared feature representation and the raw pixels of the image as feature.}
\label{tab:ImageRetrievalSharedFeatures}
\end{table}

As a baseline to compare to the retrieval image based on the shared features, we use the raw pixels of the image to find the image location. Our experiments show that using raw pixels as feature yields a poor performance to locate the patches (see Table \ref{tab:ImageRetrievalSharedFeatures}, second row). We note that even if the retrieved images look similar to the query images, they do not come from the same location. The recommended images using raw pixels are mainly based on the image color. Whenever a harvest fields is used as query image the retrieved images correspond to harvested fields as well. This is not the case when using shared features since seasonal changes are ignored in the shared representation.

\begin{figure}[b]
\begin{center}
\begin{tabular}{c c} 	
\raisebox{-0.5\totalheight}{\includegraphics[width=40mm, scale=1.0]{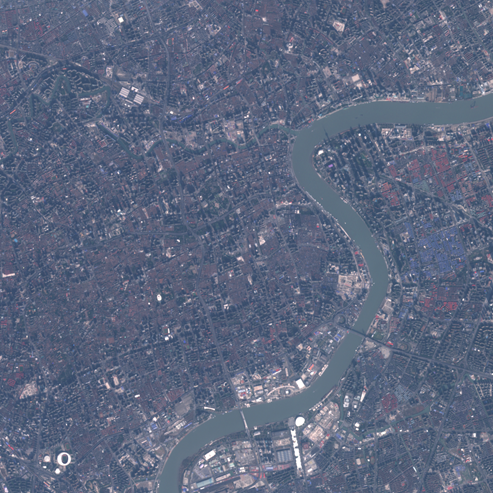}} & \raisebox{-0.5\totalheight}{\includegraphics[width=40mm, scale=1.0]{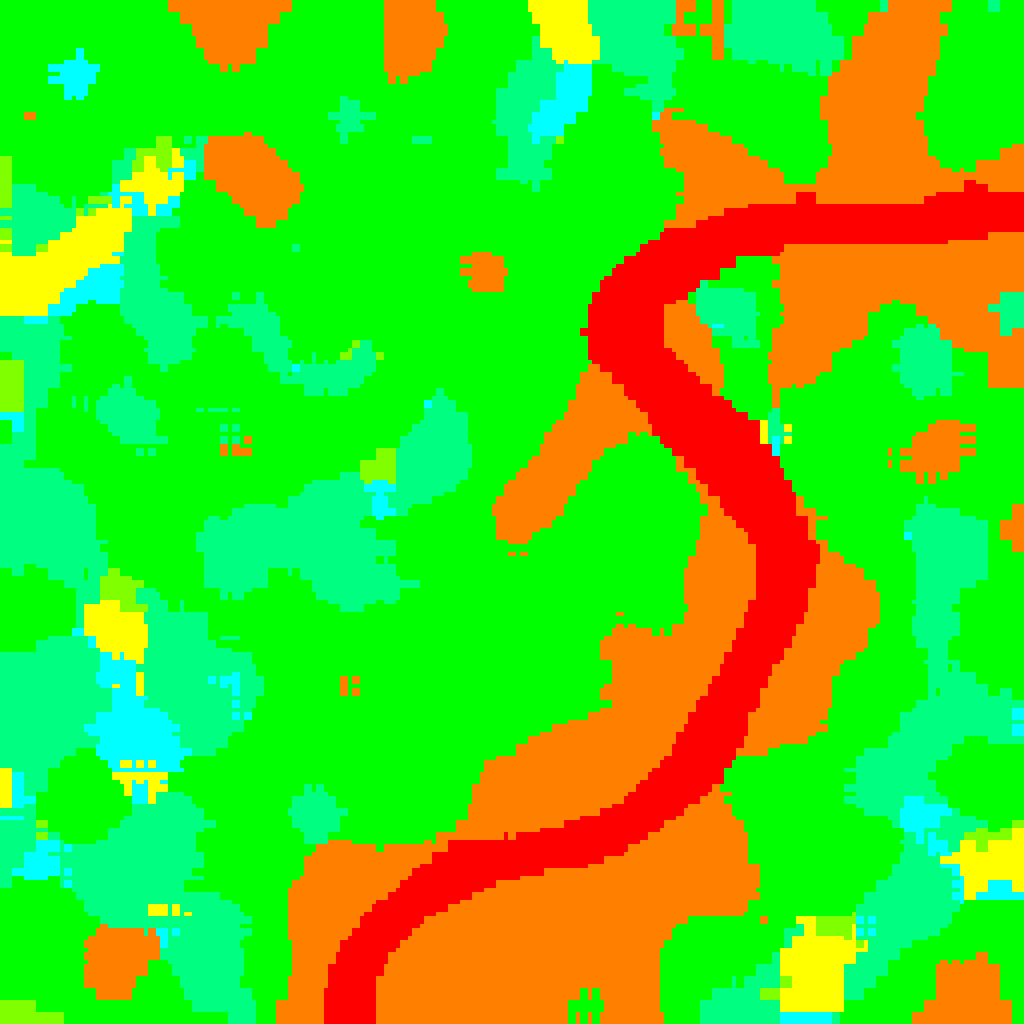}} \\[2.00cm]
(a) & (b)
\end{tabular}
\end{center}
\caption{Image segmentation in Shanghai, China. A sliding window is used to extract the shared features of the image which in turn are used to perform clustering with 7 classes. (a) Image to be segmented; (b) Segmentation map. }
\label{fig:SegmentationExperiment}
\end{figure}

\subsection{Image classification}

\begin{figure*}[!t]
\begin{center}
\begin{tabular}{c c c} 	
\raisebox{-0.5\totalheight}{\includegraphics[width=55mm, scale=1.0]{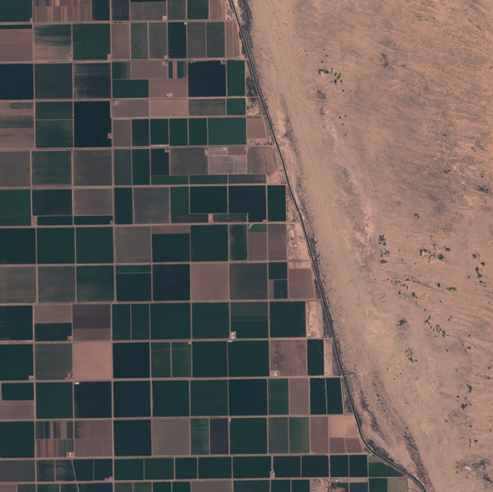}} & \raisebox{-0.5\totalheight}{\includegraphics[width=55mm, scale=1.0]{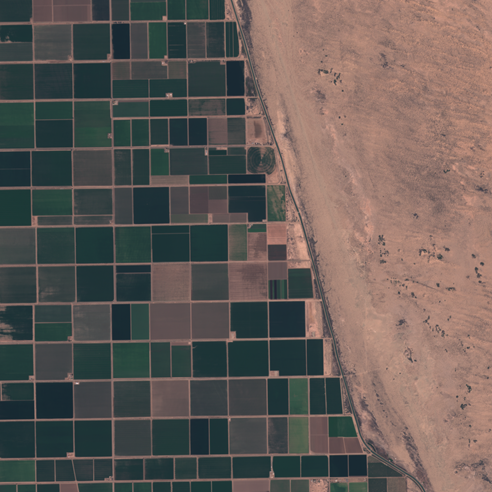}} & \raisebox{-0.5\totalheight}{\includegraphics[width=55mm, scale=1.0]{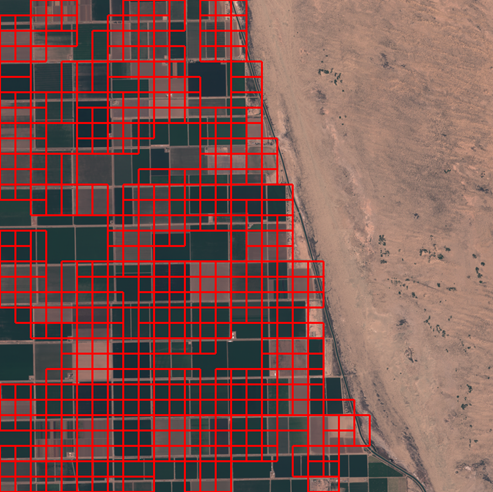}} \\[3.00cm]
(a) & (b) & (c)
\end{tabular}
\end{center}
\caption{Change detection in Brawley, USA. A window slides on the images to be compared and extracts their exclusive features. (a) Image $\mathbf{x}$; (b) Image $\mathbf{y}$; (c) A red colored box is plotted in the regions where the $L_1$ distance of the exclusive features extracted by the window is higher than a user-selected threshold.}
\label{fig:ChangeDetectionExperiment}
\end{figure*}

A common method to evaluate the performance of unsupervised features is to apply them to perform image classification. We test the shared features extracted by our model using a novel dataset called EuroSAT \cite{DBLP:journals/corr/abs-1709-00029}. It contains 27000 labeled images in 10 classes (residential area, sea, river, highway, etc.). We divide the dataset into a training and test dataset using a 80:20 split keeping a proportional number of examples per class. 

We recover the shared feature encoder $\mathbf{E}_{sh}(\cdot)$ as feature extractor from the pretrained model. We append two fully-connected layers of 64 and 10 units, respectively on top of the feature extractor. We only train these fully-connected layers while keeping frozen the weights of the feature extractor in a supervised manner using the training split of EuroSAT. Results can be observed in Table \ref{tab:ClassificationResults}. We obtain an accuracy of $92.38\%$ while we achieve an accuracy of $94.54\%$ by not freezing the weights of the feature extractor during training. It is important to note that using pretrained weights reduces the training time and allows to achieve better performance with respect to randomly initialized weights ($62.13\%$ of accuracy after 50 epochs).

\begin{table}[h]
\begin{center}
  \begin{tabular}{| c | c | c |}
    \hline
		Model & Accuracy  & Epochs\\ \hline
		\hline
    Pretrained + Fine-tuning    & $92.38\%$ & $10$\\ \hline
		Pretrained + Full-training  & $94.54\%$ & $10$\\ \hline
		From scratch                & $62.13\%$ & $50$\\
    \hline
  \end{tabular}
\end{center}
\caption{Accuracy results in the test dataset from classification experiments.}
\label{tab:ClassificationResults}
\end{table}

Higher accuracy ($98.57\%$) in the EuroSAT dataset is claimed by Helber \etal \cite{DBLP:journals/corr/abs-1709-00029} using supervised pretrained GoogLeNet or ResNet-50 models. However, we show that using a very simple model trained in an unsupervised manner allows us to obtain excellent results.

\subsection{Image segmentation}

Since the shared feature representation are related to the location and texture of the image, we perform a qualitative experiment to illustrate that it can be used to perform image segmentation. An image of size $1024 \times 1024 \times 4$ is randomly selected from a time series. Then, a sliding window of size $64 \times 64 \times 4$ and stride of size $32 \times 32$ is used to extract the patches. The shared features extracted from these patches are used to perform clustering via k-means. A new sliding window with a stride of $8 \times 8$ is used to extract the shared features from the image. Each of the extracted shared features is assigned to a cluster. Since several clusters are assigned for each pixel, the cluster is decided by the majority of voted clusters. In Figure \ref{fig:SegmentationExperiment}, a segmentation map example in Shanghai is displayed. As can be seen, this unsupervised image segmentation method achieves interesting results. It is able to segment the river, the port area and the residential area, among others. We think that segmentation results can be improved by using a smaller windows to achieve better resolution. On the other hand, experiments using the raw pixels of the image as features produce segmentation maps of lower visual quality.

\subsection{Change detection}

Another qualitative experiment is performed in order to illustrate some properties of the exclusive feature representation. Since the particular information of each image of a time series is contained in the exclusive feature, we leverage this information to propose a naive change detection method. Two images of size $1024 \times 1024 \times 4$ are selected from a given time series. A sliding window of size $64 \times 64 \times 4$ is used to explore both images using a stride of size $32 \times 32$. As the window slides, the exclusive features are extracted and compared using the $L_1$ distance. Then, a threshold is defined to determine whether a change has occurred or not. If the $L_1$ distance is higher than the threshold, a red colored box is plotted indicating that change has occurred. An example can be seen in Figure \ref{fig:ChangeDetectionExperiment}. Despite the method simplicity, our experiments suggest that the low-dimensional exclusive feature captures the factors of variation in time series generating visually coherent change detection maps. 

%
%


\section{Conclusion}

In this work, we investigate how to obtain a suitable data representation of satellite image time series. We first present a model based on VAE and GAN methods combined with the cross-domain autoencoder principle. This model is able to learn a disentangled representation that consists of a common representation for the images of the same time series and an exclusive representation for each image. We train our model using Sentinel-2 time series which indicates that the model is able to deal with huge amounts of high-dimensional data. Finally, we show experimentally that the disentangled representation can be used to achieved interesting results at multiple tasks such as image classification, image retrieval, image segmentation and change detection.


{\small
\bibliographystyle{ieee}
\bibliography{egbib}
}

\end{document}